\documentclass[11pt,a4paper]{article}

\usepackage{placeins}

\usepackage{graphicx}
\usepackage{amsmath,amssymb,amsthm} 
\usepackage{color}

\usepackage{scalerel}
\usepackage{authblk}
\usepackage{tikz}
\usetikzlibrary{svg.path,calc}

\definecolor{orcidlogocol}{HTML}{A6CE39}
\tikzset{
  blaaa/.pic={
    \fill[orcidlogocol] svg{M256,128c0,70.7-57.3,128-128,128C57.3,256,0,198.7,0,128C0,57.3,57.3,0,128,0C198.7,0,256,57.3,256,128z};
    \fill[white] svg{M86.3,186.2H70.9V79.1h15.4v48.4V186.2z}
                 svg{M108.9,79.1h41.6c39.6,0,57,28.3,57,53.6c0,27.5-21.5,53.6-56.8,53.6h-41.8V79.1z M124.3,172.4h24.5c34.9,0,42.9-26.5,42.9-39.7c0-21.5-13.7-39.7-43.7-39.7h-23.7V172.4z}
                 svg{M88.7,56.8c0,5.5-4.5,10.1-10.1,10.1c-5.6,0-10.1-4.6-10.1-10.1c0-5.6,4.5-10.1,10.1-10.1C84.2,46.7,88.7,51.3,88.7,56.8z};
  }
x}

\newcommand\orcidicon[1]{\href{https://orcid.org/#1}{\mbox{
\begin{tikzpicture}[overlay,remember picture]
\coordinate (A);
\coordinate(B) at ($(A)-(7pt,-12pt)$);
\end{tikzpicture}
\begin{tikzpicture}[overlay,remember picture,yscale=-0.03,xscale=0.03,transform shape]
\pic at (B) {blaaa};
\end{tikzpicture}
}{\hspace{-1.5ex}}}}

\usepackage[utf8]{inputenc}
\usepackage{cite}

\usepackage{mathtools}
\usepackage{esint}
\usepackage{array}
\usepackage[flushleft,alwaysadjust]{paralist}
\usepackage{bm}
\usepackage{fix-cm} 

\usepackage{colonequals}
\usepackage[active]{srcltx}
\usepackage[xspace]{ellipsis}
\usepackage{algorithmic}
\usepackage{amsmath}
\usepackage{amssymb}
\usepackage{bm}
\usepackage{booktabs}
\usepackage{calc}
\usepackage{comment}
\usepackage{epsfig}
\usepackage{flushend}
\usepackage{graphicx}
\usepackage{ifthen}
\usepackage{multirow}
\usepackage{relsize}
\usepackage{setspace}
\usepackage{siunitx}

\usepackage{caption}
\usepackage[caption=false]{subfig}
\usepackage{xspace}

\usepackage{url}

\usepackage[colorlinks,citecolor=blue,urlcolor=blue]{hyperref}

\DeclareMathOperator{\tvec}{vec}

\DeclareMathOperator{\ctwoh}{hom}
\DeclareMathOperator{\h2c}{hom^{-1}}

\theoremstyle{remark}
\newtheorem{problem}{Problem}

\newcommand{\0}{\M{0}}

\newcommand{\Man}{\mathcal{H}}

\newcommand{\Mt}[1]{\tilde{\M{#1}}}
\newcommand{\M}[1]{\mathbf{#1}}
\newcommand{\R}{\mathbb{R}}
\newcommand{\T}{\top}

\newcommand{\costML}{J_{\mathrm{ML}}}

\newcommand{\frob}[1]{\ensuremath{\| #1 \|_{\mathrm{F}}}}

\newcommand{\norm}[1]{\left\lVert#1\right\rVert}

\newcommand{\veot}[1]{\underline{\tilde{\ve{#1}}}}

\newcommand{\ve}[1]{\mathbf{#1}}
\newcommand{\vt}[1]{\tilde{\ve{#1}}}

\newcommand{\Pib}{\M{\Pi}}

\newcommand{\etb}{\bm{\eta}}

\makeatletter
\DeclareRobustCommand\onedot{\futurelet\@let@token\@onedot}
\def\@onedot{\ifx\@let@token.\else.\null\fi\xspace}

\def\etal{{et al}\onedot}

\makeatother

\newenvironment{changemargin}[2]{%
  \begin{list}{}{%
      \setlength{\topsep}{0pt}%
      \setlength{\leftmargin}{#1}%
      \setlength{\rightmargin}{#2}%
      \setlength{\listparindent}{\parindent}%
      \setlength{\itemindent}{\parindent}%
      \setlength{\parsep}{\parskip}%
    }%
  \item[]}
  {\end{list}}

\newenvironment{keywords}{%
  \begingroup
  \def\and{\unskip\space\textperiodcentered\space\ignorespaces}
  \begin{changemargin}{\leftmargin}{\leftmargin}
    \small\noindent\emph{Keywords}:}
  {\end{changemargin}
  \endgroup
}

\begin{document}

\title{Full Explicit Consistency  Constraints in Uncalibrated Multiple
  Homography Estimation}

\author{Wojciech Chojnacki \protect\orcidicon{0000-0001-7782-1956}}
  
\author{Zygmunt L. Szpak \protect\orcidicon{0000-0002-0694-4622}}

\affil{%
  Australian Institute for Machine Learning \\ The University of Adelaide, SA 5005,
  Australia}
\affil{%
  \texttt{\smaller[0.5]
  \{wojciech.chojnacki, zygmunt.szpak\}@adelaide.edu.au}
}

\date{}

\maketitle

\begin{abstract}
  We reveal a complete set of constraints that need to be imposed on a set of $3 \times 3$ matrices to ensure that the matrices represent genuine homographies associated with multiple planes between two views. We also show how to exploit the constraints to obtain more accurate estimates of homography matrices between two views.  Our study resolves a long-standing research question and provides a fresh perspective and a more in-depth understanding of the multiple homography estimation task.
\end{abstract}


\section{Introduction}
\label{sec:introduction}

Two images of the same planar surface in space are related by a homography---a transformation which can be described, to within a scale factor, by an invertible $3 \times 3$ matrix.  This basic fact is what makes estimating a \emph{single} homography from image measurements one of the primary tasks in computer vision.  Three-dimensional reconstruction, mosaicing, camera calibration, and metric rectification are examples of the applications making use of a single homography~\cite{Hartley2004Multiple}.  A recent addition to this list is the problem of color transfer \cite{finlayson17:_color_homog,gong16:_recod}.  Various methods for estimating a single homography are available~\cite{Hartley2004Multiple} and new techniques emerge on a regular basis \cite{osuna-enciso16:_multiob_approac_homog_estim,zhao16:_accur_half_sift, qi17:_fast,barath17, guo17}.

A task closely related to estimating a single homography is that of estimating \emph{multiple} homographies.  Multiple planar surfaces are ubiquitous in urban environments, and, as a result, estimating multiple homographies between two views from image measurements is an important step in many applications such as non-rigid motion detection~\cite{kahler:_rigid,zelnik02}, enhanced image warping~\cite{gao11:_const}, multiview 3D reconstruction~\cite{karami14:_multiv}, augmented reality~\cite{prince02:_augmen}, indoor navigation~\cite{rodrigo09:_robus_effic_featur_track_indoor_navig}, multi-camera calibration~\cite{ueshiba03:_plane}, camera-projector calibration~\cite{park10:_activ_calib_camer_projec_system}, or ground-plane recognition for object detection and tracking~\cite{arrospide10:_homog}.  Surprising as it may seem, a vast array of techniques for estimating multiple homographies, including many robust multi-structure estimation methods~\cite{chin:robust_stat_apprach2009,chin:ork2009,isack12:_energy_geometric_fitting,pham14:_rand_clust_model, wang12:_sim_fitting_multi_struct,mittal12:_gener_projec_based_m_estim, decrouez:_extracting_planar_struc,fouhey:jlinkageplanes,wong:rmf,zuliani05_multiRansac} applicable to the task of estimating multiple homographies, are deficient in a fundamental way---they fail to recognise that a set of homography matrices does not represent a set of genuine homographies between two views of the same scene unless appropriate \emph{consistency constraints} are satisfied.  These constraints reflect the rigidity of the motion and the scene.  If the constraints are not deliberately enforced, they do not hold in typical scenarios.  Hence, one of the fundamental problems in estimating multiple homography matrices is to find a way to enforce the consistency constraints---a task reminiscent of that of enforcing the rank-two constraint in the case of the fundamental matrix estimation \cite[Sect. 11.1.1]{Hartley2004Multiple}.

Being unable to specify explicit formulae for all relevant constraints, various researchers have managed over the years to identify and enforce various reduced sets of constraints.  As pioneers in this regard, Shashua and Avidan~\cite{shashua96} found that homography matrices induced by four or more planes in a 3D scene appearing in two views span a four-dimensional linear subspace.  Chen and Suter~\cite{chen09:_rank_const_for_homog_over_two_views} derived a set of strengthened constraints for the case of three or more homographies in two views.  Zelnik-Manor and Irani~\cite{zelnik02} have shown that another rank-four constraint applies to a set of so-called relative homographies generated by two planes in four or more views. These latter authors also derived constraints for larger sets of homographies and views.  Finally, in recent work \cite{szpak15:_robus} Szpak \etal introduced what they dubbed the multiplicity and singularity constraints that apply to two or more, and three or more, homographies between two views, respectively.

Once isolated, the available constraints are typically put to use in a procedure whereby first individual homography matrices are estimated from image data, and then the resulting estimates are upgraded to matrices satisfying the constraints.  Following this pattern, Shashua and Avidan as well as Zelnik-Manor and Irani used low-rank approximation under the Frobenius norm to enforce the rank-four constraint.  Chen and Suter enforced their set of constraints also via low-rank approximation, but then employed the Mahalanobis norm with covariances of the input homographies.  All of these estimation procedures produce matrices that satisfy only incomplete constraints so their true consistency cannot be guaranteed.

Without knowledge of \emph{explicit} formulae for all of the constraints, it is still possible to \emph{implicitly} enforce full consistency by exploiting a natural parametrisation of the family of all fixed-size sets of compatible homography matrices (see Section~\ref{sec:path-constraints}).  Following this path, Chojnacki \etal \cite{chojnacki:_multi_homgr_full_consist,chojnacki15:_enfor} employed this parametrisation and a distinct cost function to develop an upgrade procedure based on unconstrained optimisation.  Szpak \etal \cite{szpak14:_samps} used the same parametrisation and the Sampson distance to develop an alternative estimation technique with a sound statistical basis.  The parameters encoding compatible homographies constitute the \emph{latent variables} in the model explaining the dependencies between the homographies involved.  While the use of latent variables guarantees the enforcement of all of the underlying consistency constraints, it also has some notable drawbacks.  Specifically, the latent variable based method does not provide a means to directly \emph{measure} the extent to which a collection of homography matrices are compatible.  Furthermore, finding suitable initial values for the latent variables is a non-trivial task.  The initialisation methods utilised by Chojnacki \etal \cite{chojnacki:_multi_homgr_full_consist,chojnacki15:_enfor} and Szpak \etal \cite{szpak14:_samps} are based on factorising a collection of homography matrices. The factorisation procedure is described in detail in \cite[Sect.  6.2]{chojnacki15:_enfor} and summarised in \cite[Algorithm 1]{chojnacki15:_enfor}. It involves a series of algebraic manipulations and a singular value decomposition. Each of these steps is sensitive to noise, and so when some of the given homographies have substantial uncertainty, the resulting initial latent variables will correspond to compatible homographies with high reprojection error. The high reprojection means that the subsequent optimisation process could converge to a sub-optimal local minimum. This predicament is explained and illustrated in Figure 4 of \cite{szpak14:_samps}. 

\begin{figure}[t]
  \centering
  \subfloat[\label{fig:1}]{%
    \includegraphics[width=0.49\textwidth]{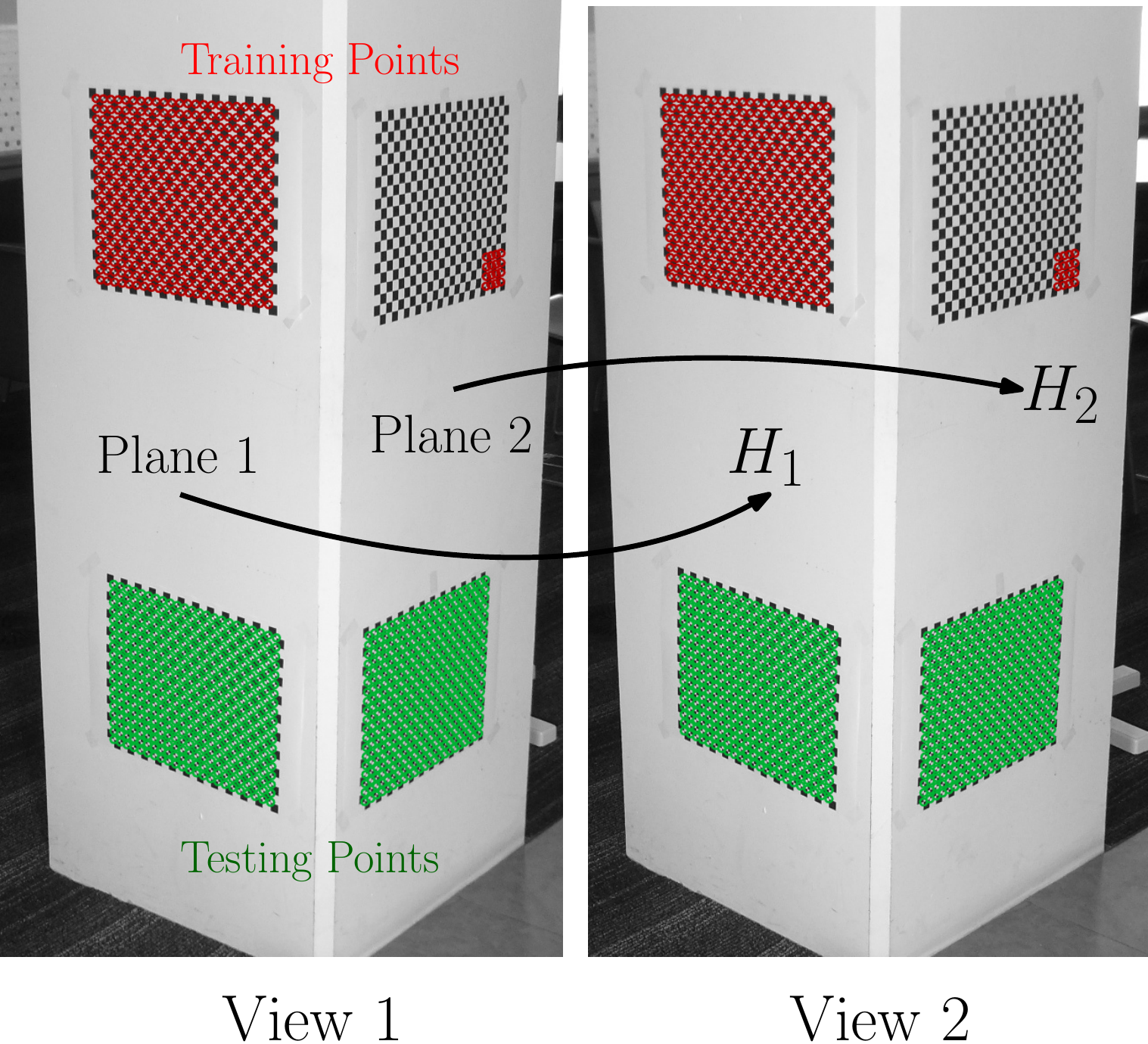} 
  } 
 \hfill 
  \subfloat[ \label{fig:2}]{%
    \includegraphics[width=0.238\textwidth]{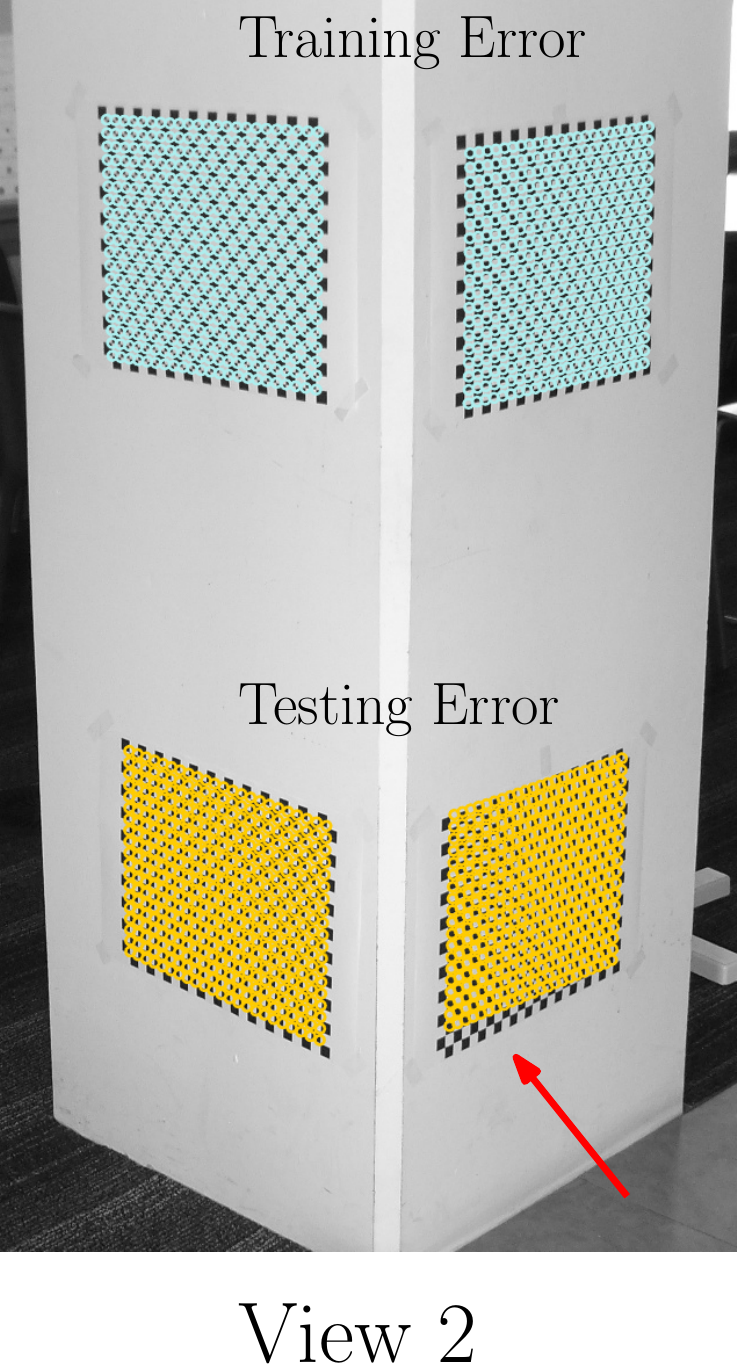} 
  } 
  \hfill 
  \subfloat[\label{fig:3}]{%
    \includegraphics[width=0.238\textwidth]{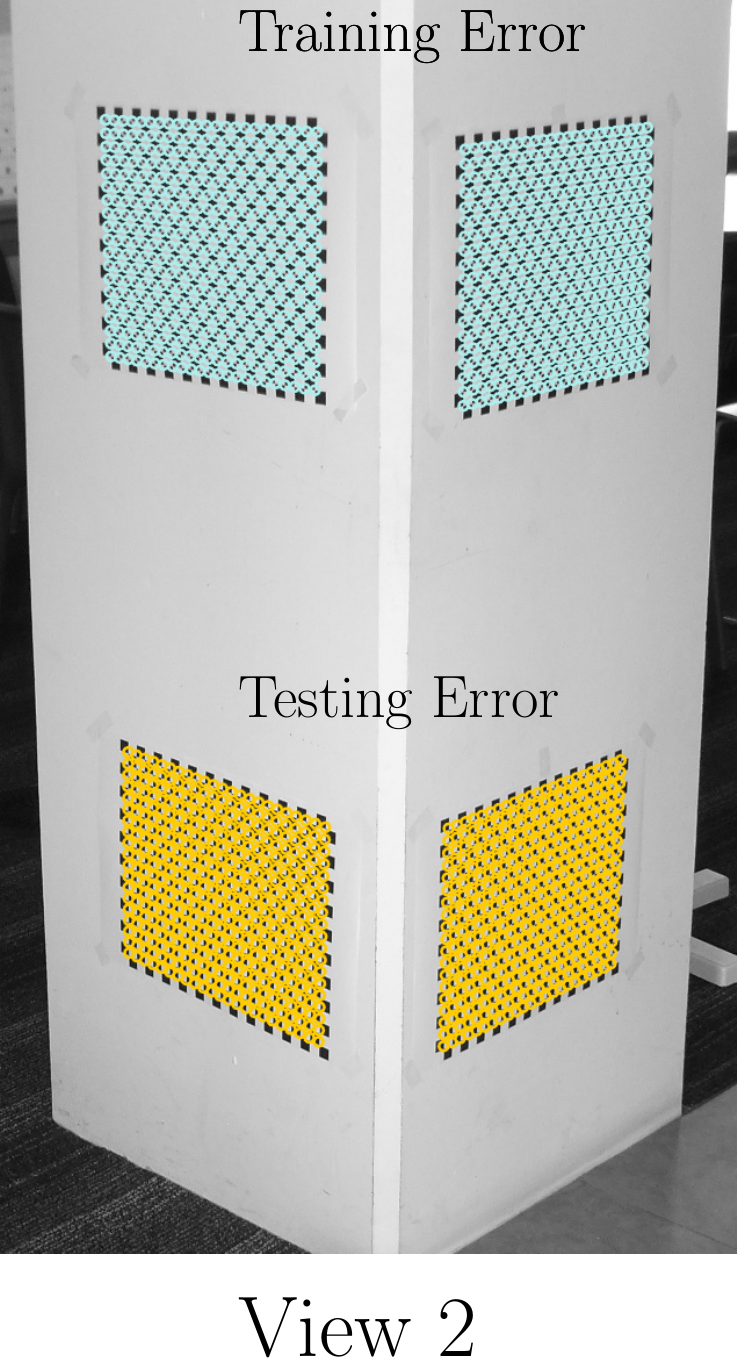} 
  } 
  \caption{Enforcing full consistency constraints improves the quality of homography estimates. Panel (a) illustrates red feature points on two planes which were used to estimate two homography matrices ($H_1$ and $H_2$). The quality of the estimated homographies can be evaluated by how accurately they map the green feature points into the second view.  Panel (b) shows the result of mapping points in the first view to the second view using homographies estimated using the \emph{gold standard} bundle-adjustment method which does not enforce homography consistency constraints. Panel (c) illustrates the result of mapping points in the first view to the second view using homographies estimated using bundle-adjustment while simultaneously enforcing the specific constraints proposed in this paper. Note how the gold standard method fails to map the green points associated with the second plane accurately; this is indicated by the red arrow. In contrast, our proposed solution produces a substantially more accurate result. }
\label{fig:benefits-exp-constr}
\end{figure}

In this paper we exhibit a full set of explicit constraints for multiple homographies between two views. This constitutes a theoretical contribution and also has practical ramifications.  We use the deduced set of constraints to define a quantifiable measure to assess the extent to which separately estimated homographies are mutually incompatible.  Based on this measure, we demonstrate experimentally that unless the consistency constraints are explicitly enforced, estimates of multiple homographies cannot be treated as \emph{bona fide} homographies between two views. The palpable advantage of our constrained homography estimation procedure is evident in Figure~\ref{fig:benefits-exp-constr}.   By imposing consistency constraints, one improves not only the accuracy of the homographies but also ensures that any derived quantities, e.g.\ camera projection matrices, will be more accurate.

\section{Path to Constraints}
\label{sec:path-constraints}


As already pointed out in the introduction, when estimating a set of homographies associated with multiple planes from image correspondences between two views, one must recognise that the homographies involved are interdependent.  To reveal the nature of the underlying dependencies, consider two fixed uncalibrated cameras giving rise to two camera matrices $\M{P}_1 = \M{K}_{1}\M{R}_{1}[\M{I}_{3}, -\ve{t}_{1}]$ and $\M{P}_2 = \M{K}_{2}\M{R}_{2}[\M{I}_{3}, -\ve{t}_{2} ]$.  Here, the length-3 translation vector $\ve{t}_{k}$ and the $3 \times 3$ rotation matrix $\M{R}_{k}$ represent the Euclidean transformation between the $k$-th ($k = 1,2$) camera and the world coordinate system, $\M{K}_{k}$ is a $3 \times 3$ upper triangular calibration matrix encoding the internal parameters of the $k$-th camera, and $\M{I}_3$ denotes the $3 \times 3$ identity matrix.  Suppose, moreover, that a set of $I$ planes in a 3D scene have been selected.  Given $i = 1, \dots, I$, let the $i$-th plane from the collection have a unit outward normal $\ve{n}_i$ and be situated at a distance $d_i$ from the origin of the world coordinate system. Then, for each $i = 1, \dots, I$, the $i$-th plane gives rise to a planar homography between the first and second views described by the $3 \times 3$ matrix
\begin{equation}
  \label{eq:1}
  \M{H}_i = w_i \M{A} + \ve{b} \ve{v}_i^{\T},
\end{equation}
where
\begin{equation}
  \label{eq:2}
  \begin{aligned}
    \M{A} & =\M{K}_{2}\M{R}_{2}\M{R}_{1}^{-1}\M{K}_{1}^{-1}, & 
    w_{i} & = d_{i} - \ve{n}_{i}^{\T}\ve{t}_{1}, \\
    \ve{b} & =\M{K}_{2}\M{R}_{2}(\ve{t}_{1} - \ve{t}_{2}), &
    \ve{v}_{i} & =\M{K}^{-\T}_{1} \M{R}_{1} \ve{n}_{i}
  \end{aligned}
\end{equation}%
(cf.\ \cite{faugeras88:_motion_struc_motion_planar_envir,szpak13:_const}).  In the case of calibrated cameras when one may assume that $\M{K}_1 = \M{K}_2 = \M{I}_3$, $\ve{t}_1 = \0$, $\M{R}_1 = \M{I}_3$, $\M{R}_2 = \M{R}$, system \eqref{eq:2} reduces to
\begin{equation}
  \label{eq:3}
  \begin{aligned}
    \M{A} & =\M{R}, & 
    w_{i} & = d_{i}, \\
    \ve{b} & = \ve{t}, &
    \ve{v}_{i} & = \ve{n}_{i},
  \end{aligned}
\end{equation}
with $\ve{t} = - \M{R}\ve{t}_2$, and equality \eqref{eq:1} becomes the familiar \emph{direct nRt} representation
\begin{math}
  \M{H}_i = d_i \M{R} + \ve{t} \ve{n}_i^\T
\end{math}
(cf.\ \cite{baker:_param_homog}, \cite[Sect. 5.3.1]{ma05:_invit_d_vision}).  We stress that all of our subsequent analysis concerns the general uncalibrated case, with $\M{A}$, $\ve{b}$, $w_i$'s and $\ve{v}_i$'s to be interpreted according to \eqref{eq:2} rather than~\eqref{eq:3}.

An entity naturally associated with the matrices $\M{H}_i$ is the $3 \times 3I$ (horizontal) concatenation matrix
\begin{math}
  \M{H}_{\mathrm{cat}} = [\M{H}_1, \dots, \M{H}_I].
\end{math}
With $\tvec$ denoting column-wise vectorisation \cite{luetkepol96:_handb_matric}, if we let
\begin{displaymath}
  \etb = [(\tvec \M{A})^\T,  \ve{b}^\T, \ve{v}_1^\T, \dots,
  \ve{v}_I^\T, w_1, \dots, \linebreak[0] w_I]^\T
\end{displaymath}
and
\begin{displaymath}
  \Pib(\etb) = [\Pib_1(\etb), \dots, \Pib_I(\etb)],
\end{displaymath}
where
\begin{equation}
  \label{eq:4}
  \Pib_i(\etb) = w_i \M{A} + \ve{b} \ve{v}_i^{\T}
  \quad
  (i = 1, \dots, I),
\end{equation}
then $\M{H} _{\mathrm{cat}}$ can be written as
\begin{displaymath}
  \M{H} _{\mathrm{cat}} = \Pib(\etb).
\end{displaymath}
Here $\etb$ represents a vector of latent variables that link all the constituent matrices together and provide a natural parametrisation of the set $\Man$ of all $ \M{H}_{\mathrm{cat}}$'s.  Since $\etb$ has a total of $4I + 12$ entries, the aggregate of all matrices of the form $\Pib(\etb)$ has dimension no greater than $4I + 12$, with the relevant notion of dimension being here that of dimension of a semi-algebraic set \cite{chojnacki11:_dim_res,chojnacki12}.  By employing a rather subtle argument, one can calculate exactly the dimension of the set of all $\Pib(\etb)$'s (which is the same as the dimension of $\Man$ given that the set of all $\Pib(\etb)$'s is identical with $\Man$) and this turns out to be equal to $4I + 7$ \cite{chojnacki11:_dim_res,chojnacki12}.  The difference between the dimension of the set of all $\etb$'s and the dimension of $\Man$ is indicative of five degrees of the internal gauge freedom present in the parametrisation $\Pib$; this occurrence will be crucially exploited in what follows---see Section~\ref{sec:problem}.  Since $4I + 7 < 9I$ ($= 3 \times 3I$) whenever $I \geq 2$, it follows that $\Man$ is a proper subset of the set of all $3 \times 3I$ matrices for $I \geq 2$.  It is now clear that the requirement that $\M{H} _{\mathrm{cat}}$ take the form as per \eqref{eq:4} whenever $I \geq 2$ can be seen as an implicit constraint on $\M{H} _{\mathrm{cat}}$, with the consequence that the $\M{H}_i$'s are all interdependent.  This further begs the question as to how to turn the implicit constraint into a system of explicit constraints (not involving the latent variables) that has to be put upon a set of matrices $\M{H}_i$ ($i = 1, \dots, I$) in order that the $\M{H}_i$'s represent genuine homographies between two views.  We shall subsequently answer this question in steps, with the first step being taken in the next section.

\section{Problem}
\label{sec:problem}

Let $\R$ denote the set of real numbers and let $\R^{m \times n}$ denote the set of $m \times n$ matrices with entries in $\R$.  We formally formulate the main purpose of this paper as an answer to the following problem:

\begin{problem}
  \label{prob:1}
  Given $I$ invertible matrices $\M{H}_1$, \dots, $\M{H}_I \in \R^{3 \times 3}$, find a system of equations that the $\M{H}_i$'s have to satisfy in order to be representable in the form
  \begin{equation}
    \label{eq:5}
    \M{H}_i  =  w_i \M{A} + \ve{b} \ve{v}_i^\T
    \quad
    (i = 1, \dots, I)
  \end{equation}
  for some matrix $\M{A} \in \R^{3 \times 3}$, some vectors $\ve{b}$, $\ve{v}_1$, \dots, $\ve{v}_I \in \R^3$, and some scalars $w_1$, \dots, $w_I \in \R$.
\end{problem}

We start with two observations that will greatly facilitate solving the problem.  First we note that, for each $i$, if $\M{H}_i$ can be represented as $ w_i \M{A} + \ve{b} \ve{v}_i^\T$, then necessarily $w_i \neq 0$.  Indeed, if $w_i = 0$ held for some $i$, then $\M{H}_i$ would be equal to $\ve{b} \ve{v}_i^\T$ and hence would be of rank one, contravening the assumption that all the $\M{H}_i$'s are invertible.  Next we observe that if \eqref{eq:5} holds for a set of $\M{A}$, $\ve{b}$, $\ve{v}_i$'s, and $w_i$'s, then it also holds for various other sets of $\M{A}$, $\ve{b}$, $\ve{v}_i$'s, and $w_i$'s.  Indeed, if
\begin{math}
  \M{H}_i = w_i \M{A} + \ve{b} \ve{v}_i^\T
\end{math}
for each $i$, then also
\begin{math}
  \M{H}_i = w_i' \M{A}' + \ve{b}' \ve{v}_i'^\T
\end{math}
for each $i$, where $\M{A}' = \beta \M{A} + \ve{b} \ve{c}^\T$, $\ve{b}' = \alpha \ve{b}$, $\ve{v}_i' = \alpha^{-1} \ve{v}_i - \alpha^{-1} \beta^{-1} w_i \ve{c}$, and $w_i' = \beta^{-1} w_i$, with $\alpha$ and $\beta$ being non-zero scalars and $\ve{c}$ being a length-$3$ vector.  We now exploit this last observation by letting $\alpha = 1$, $\beta = w_1$, and $\ve{c} = \ve{v}_1$; critically, by our first observation, $\beta$ is non-zero.  Then $\M{A}'$ becomes $\M{H}_1$ and we further have
\begin{math}
  \M{H}_i = w_i' \M{H}_1 + \ve{b} \ve{v}_i'^\T
\end{math}
with $w_i' = w_1^{-1}w_i$ and $\ve{v}_i' = \ve{v}_i - w_1^{-1} w_i \ve{v}_1$ for each $i$.  In light of this, we see that Problem~\ref{prob:1} can equivalently be restated as follows:

\begin{problem}
  \label{prob:2}
  Given $I$ invertible matrices $\M{H}_1$, \dots, $\M{H}_I \in \R^{3 \times 3}$, find a system of equations that the $\M{H}_i$'s have to satisfy in order that
  \begin{equation}
    \label{eq:6}
    \M{H}_i = w_i \M{H}_1 + \ve{b} \ve{v}_i^\T
    \quad
    (i = 2, \dots, I)
  \end{equation}
  hold for some vectors $\ve{b}$, $\ve{v}_2$, \dots, $\ve{v}_I \in \R^3$ and some scalars $w_2$, \dots, $w_I \in \R$.
\end{problem}

In what follows we reveal a solution to Problem~\ref{prob:2}. This will give us a sought-after set of explicit homography constraints.

\section{Algebraic Prerequisites}
\label{sec:algebr-prer}

To make the derivation of a constraint set more accessible, we start in this section with some necessary technical prerequisites.

\subsection{The Characteristic Polynomial}
\label{sec:char-polyn}

Let $\M{A}, \M{B} \in \R^{3 \times 3}$.  The \emph{linear matrix pencil} of the matrix pair $(\M{A}, \M{B})$ is the matrix function $\lambda \mapsto \M{A} - \lambda \M{B}$.  The \emph{characteristic polynomial} of $(\M{A}, \M{B})$, $p_{\M{A}, \M{B}}$, is defined by
\begin{math}
  p_{\M{A}, \M{B}}(\lambda) = \det(\M{A} - \lambda \M{B}).
\end{math}
Adopting MATLAB's notation to let $\M{M}_{:i}$ represent the $i$th column of the matrix $\M{M}$, one verifies directly that $p_{\M{A}, \M{B}}$ can be explicitly written as
\begin{math}
  p_{\M{A},\M{B}}(\lambda) = \sum_{n=0}^3 (-1)^n c_n \lambda^n,
\end{math}
where
\begin{equation}
\label{eq:7}
\begin{aligned}
  c_0 & =  \det \M{A},
  \\
  c_1 & = \det [\M{B}_{:1},  \M{A}_{:2},  \M{A}_{:3}]
        +
        \det [\M{A}_{:1}, \M{B}_{:2}, \M{A}_{:3}]
  +
        \det [\M{A}_{:1}, \M{A}_{:2}, \M{B}_{:3}],
  \\
  c_2 & = \det [\M{A}_{:1},  \M{B}_{:2}, \M{B}_{:3}]
        + \det [\M{B}_{:1}, \M{A}_{:2}, \M{B}_{:3}]
  +
        \det [\M{B}_{:1}, \M{B}_{:2}, \M{A}_{:3}],
  \\
  c_3 & = \det \M{B}.
\end{aligned}
\end{equation}
%
%
%
%
The characteristic polynomial arises in connection with the generalised eigenvalue problem
\begin{equation}
  \label{eq:8}
  \M{A} \ve{x} = \lambda \M{B} \ve{x}.
\end{equation}
As with the standard eigenvalue problem, eigenvalues for the problem \eqref{eq:8} occur precisely where the matrix pencil $\lambda \to \M{A} - \lambda \M{B}$ is singular. In other words, the eigenvalues for the pair $(\M{A}, \M{B})$ are the roots of $p_{\M{A}, \M{B}}$.

A fact that will be of significance in what follows is that if the generalised eigenvalue problem \eqref{eq:8} has a double eigenvalue, then this eigenvalue is a double root of $p_{\M{A},\M{B}}$.  For the sake of completeness, we recall the argument presented in  \cite{szpak15:_robus} which  validates this fact and correct a misprint that has slipped into the original proof.

Suppose that the generalised eigenvalue problem \eqref{eq:8} has a double eigenvalue $\mu$, which means that there exist linearly independent length-$3$ vectors $\ve{v}_1$ and $\ve{v}_2$ such that $\M{A} \ve{v}_i = \mu \M{B} \ve{v}_i$ for $i = 1,2$. With a view to showing that $\mu$ is a double root of $p_{\M{A}, \M{B}}$, select arbitrarily a length-$3$ vector $\ve{v}_3$ that does not belong to the linear span of $\ve{v}_1$ and $\ve{v}_2$; for example, we may assume that $\ve{v}_3 = \ve{v}_1 \times \ve{v}_2$.  Then $\ve{v}_1$, $\ve{v}_2$, and $\ve{v}_3$ form a basis for $\R^3$, and hence the matrix
\begin{math}
  \M{S} = [\ve{v}_1, \ve{v}_2, \ve{v}_3]
\end{math}
is non-singular.  Let
\begin{math}
  \Mt{A} = \M{S}^{-1} \M{A} \M{S}
\end{math}
and
\begin{math}
  \Mt{B} = \M{S}^{-1} \M{B} \M{S}.
\end{math}
For $i = 1, 2, 3$, let $\ve{j}_i$ denote the $i$-th standard unit vector in $\R^3$, with $1$ in the $i$-th position and $0$ in all others.  Then, clearly, $\ve{v}_i = \M{S} \ve{j}_i$ for $i = 1,2,3$.  It is immediate that, for $i = 1,2$, $\Mt{A} \ve{j}_i = \mu \Mt{B} \ve{j}_i$ and so
\begin{math}
  (\Mt{A} - \lambda \Mt{B}) \ve{j}_i
  = (\mu - \lambda) \Mt{B} \ve{j}_i.
\end{math}
Hence the pencil $\Mt{A} - \lambda \Mt{B}$ takes the form
\begin{displaymath}
  \Mt{A}  - \lambda \Mt{B} =
  \begin{bmatrix}
    (\mu - \lambda) \tilde b_{11}
    &  (\mu - \lambda) \tilde b_{12}
    & \tilde a_{13} - \lambda \tilde b_{13}
    \\
    (\mu - \lambda) \tilde b_{21}
    &  (\mu - \lambda) \tilde b_{22}
    & \tilde a_{23} - \lambda \tilde b_{23}
    \\
    (\mu - \lambda) \tilde b_{31}
    &  (\mu - \lambda) \tilde b_{32}
    & \tilde a_{33} - \lambda \tilde b_{33}
  \end{bmatrix}
\end{displaymath}
and we have
\begin{align*}
  p_{\Mt{A}, \Mt{B}}(\lambda)
  & =
  \begin{vmatrix}
    (\mu - \lambda) \tilde b_{11}
    &  (\mu - \lambda) \tilde b_{12}
    & \tilde a_{13} - \lambda \tilde b_{13}
    \\
    (\mu - \lambda) \tilde b_{21}
    &  (\mu - \lambda) \tilde b_{22}
    & \tilde a_{23} - \lambda \tilde b_{23}
    \\
    (\mu - \lambda) \tilde b_{31}
    &  (\mu - \lambda) \tilde b_{32}
    & \tilde a_{33} - \lambda \tilde b_{33}
  \end{vmatrix}
  \\
  &
    = (\mu - \lambda)^2
  \begin{vmatrix}
    \tilde b_{11}  &  \tilde b_{12}
    &  \tilde a_{13} - \lambda \tilde b_{13}
    \\
    \tilde b_{21}  &  \tilde b_{22}
    & \tilde a_{23} - \lambda \tilde b_{23}
    \\
    \tilde b_{31}  &  \tilde b_{32}
    & \tilde a_{33} - \lambda \tilde b_{33}
  \end{vmatrix},
\end{align*}
which shows that $\mu$ is a double root of $p_{\Mt{A},\Mt{B}}$.  But $p_{\Mt{A},\Mt{B}}$ coincides with $p_{\M{A},\M{B}}$, given that
\begin{align*}
  p_{\Mt{A},\Mt{B}}(\lambda)
  &  = 
  \det (\M{S}^{-1} (\M{A} - \lambda \M{B}_3) \M{S}) \\
  & =
  \det \M{S}^{-1} \det(\M{A} - \lambda \M{B}_3) \det \M{S} \\
  & = (\det \M{S})^{-1} \det(\M{A} - \lambda \M{B}_3) \det \M{S}
  = p_{\M{A},\M{B}}(\lambda).
\end{align*}
Therefore $\mu$ is \emph{a fortiori} a double root of $p_{\M{A},\M{B}}$.

\subsection{A Double Root of the  Characteristic Polynomial of a Cubic
  Polynomial}
\label{sec:double-root-char}

Let $\M{A}$ and $\M{B}$ be two $3 \times 3$ matrices such that $p_{\M{A},\M{B}}$ has a double root which is not a triple root. Then, as it turns out, the root is uniquely determined and is given by an explicit formula. This is a consequence of a more general result that we present next.

Let 
\begin{math}
  p(\lambda) = a\lambda^3  + b\lambda^2 + c\lambda + d
\end{math}
be a cubic polynomial with $a \neq 0$.  Suppose that $\mu$ is a double root of $p$,
\begin{equation}
  \label{eq:9}
  p(\mu) =  p'(\mu) = 0,
\end{equation}
but not a triple root, $p''(\mu) \neq 0$; we shall term such a double root non-degenerate.  Then equations \eqref{eq:9} can explicitly be written as
\begin{align}
  a \mu^3 + b \mu^2 + c \mu + d & = 0,
                                              \label{eq:10}
  \\
  3a \mu^2 + 2b \mu + c & = 0.
                                  \label{eq:11}
\end{align}
When we multiply the first of these equations by $3$ and the second by $\mu$ and next subtract the second equation from the first, we get
\begin{equation}
  \label{eq:12}
  b \mu^2 + 2c \mu + 3d = 0.
\end{equation}
Restating \eqref{eq:10} and \eqref{eq:12} as
\begin{align*}
  3a \mu^2 + 2b \mu
  & = -c,
  \\
  b \mu^2 + 2c \mu
  & = -3d,
\end{align*}
we obtain a system of linear equations in $\mu^2$ and $\mu$. Solving for $\mu$ and $\mu^2$ gives
\begin{equation}
  \label{eq:15}
  \mu
    = \frac{9ad - bc}{2(b^2 - 3ac)}
    \quad
    \text{and}
    \quad
  \mu^2
    = \frac{c^2 - 3bd}{b^2 - 3ac}.
\end{equation}
Here $b^2 \neq 3ac$ for otherwise equation \eqref{eq:11} would have its quadratic discriminant $4(b^2 - 3ac)$ equal to zero, with the consequence that $\mu$ would be a repeated root for $p'$ and hence a triple root for $p$.  Now, the first equation in \eqref{eq:15} provides a formula for a non-degenerate double root of a cubic polynomial.  We see in particular that if a cubic polynomial has a non-degenerate double root, then this root is uniquely determined.

In light of the above discussion it is clear that if $p_{\M{A},\M{B}}$ has a non-degenerate double root, then the root is unique, and when we denote this root by $\mu_{\M{A}, \M{B}}$, we have
\begin{math}
  \mu_{\M{A}, \M{B}} = \omega_{\M{A}, \M{B}}, 
\end{math}
where, with the notation from \eqref{eq:7},
\begin{equation}
  \label{eq:16}
  \omega_{\M{A}, \M{B}}
  =
  \frac{c_1 c_2 - 9 c_0 c_3}{2(c_2^2 - 3 c_1 c_3)}.
\end{equation}

\section{Full Constraints}
\label{sec:full-constraints}

Here we finally present a solution to Problem~\ref{prob:2} (and hence also to Problem~\ref{prob:1}).

Let $\M{H}_1, \dots, \M{H}_I \in \R^{3 \times 3}$ be such that \eqref{eq:6} holds for some $\ve{b}$, $\ve{v}_2$, \dots, $\ve{v}_I \in \R^3$ and $w_2$, \dots, $w_I \in \R$.  Fix $i \in \{2, \dots, I \}$ arbitrarily.  If $\ve{c}$ is a length-$3$ vector orthogonal to $\ve{v}_i$, then
\begin{displaymath}
  \M{H}_i \ve{c}
  =  w_i \M{H}_1 \ve{c} + \ve{b} \ve{v}_i^\T \ve{c}
  =  w_i \M{H}_1 \ve{c},
\end{displaymath}
showing that $(w_i, \ve{c})$ is an eigenpair for the pair $(\M{H}_i, \M{H}_1)$.  Since length-$3$ vectors orthogonal to $\ve{v}_i$ form a two-dimensional linear space, it follows that $w_i$ is in fact a double eigenvalue for $(\M{H}_i, \M{H}_1)$.  Using the material from Section~\ref{sec:algebr-prer} and assuming that all double roots of intervening characteristic polynomials are non-degenerate (which is generically true), we conclude that, for each $i = 2, \dots, I$, $w_i$ is uniquely defined, namely
\begin{math}
  w_i = \omega_{\M{H}_i, \M{H}_1}
\end{math}
(recall the definition given in \eqref{eq:16}).  For each $i = 2, \dots, I$, let
\begin{align}
  \label{eq:18}
  \M{J}_i & = \M{H}_i -  \omega_{\M{H}_i, \M{H}_1} \M{H}_1, \\
\intertext{and let}
  \label{eq:19}
  \M{J} & = [\M{J}_2, \dots, \M{J}_I].
\end{align}
In view of \eqref{eq:6},  
\begin{math}
  \M{J}_i = \ve{b} \ve{v}_i^\T
\end{math}
for each $i = 2, \dots, I$. Hence, letting
\begin{math}
  \ve{w} = [\ve{v}_2^\T, \dots, \ve{v}_I^\T]^\T,
\end{math}
we have
\begin{displaymath}
  \M{J} = [\ve{b} \ve{v}_2^\T, \dots, \ve{b} \ve{v}_I^\T]
        = \ve{b} [\ve{v}_2^\T, \dots, \ve{v}_I^\T]
        = \ve{b} \ve{w}^\T,
\end{displaymath}
which implies that $\M{J}$ has rank one. 

Conversely, if $\M{J}$ has rank one, then
\begin{math}
  \M{J} = \ve{b} \ve{w}^\T
\end{math}
for some length-$3$ vector $\ve{b}$ and some length-$3(I-1)$ vector $\ve{w}$ which, as any vector of this length, can be represented as
\begin{math}
  \ve{w} = [\ve{v}_2^\T, \dots, \ve{v}_I^\T]^\T
\end{math}
for some length-$3$ vectors $\ve{v}_2$, \dots, $\ve{v}_I$.  This, in conjunction with the definitions \eqref{eq:18} and \eqref{eq:19}, leads to
\begin{math}
  \M{H}_i = \omega_{\M{H}_i, \M{H}_1} \M{H}_1 + \ve{b} \ve{v}_i^\T
\end{math}
for each $i = 2, \dots, I$, which is a representation of the form required in Problem~\ref{prob:2}.

In light of the above, we see that Problem~\ref{prob:2} reduces to finding the requirement in algebraic form that $\M{J}$ have rank one.  As is well known, the relevant condition is that all $2 \times 2$ minors of $\M{J}$ should vanish \cite[\S V.2.2, Thm. 3]{mostowski64:_introd}.  To express this condition explicitly, we introduce some notation.  Given an $m \times n$ matrix $\M{A}$ and positive integers $a_1$, \dots, $a_k$ with $1 \leq a_1 < a_2 < \dots < a_{k-1} < a_k \leq m$ and positive integers $b_1$, \dots, $b_l$ with $1 \leq b_1 < b_2 < \dots < b_{l - 1} < b_l \leq n$, let
\begin{math}
  \M{A}(a_1, a_2, \dots, a_{k-1}, a_k; b_1, b_2, \dots, b_{l-1}, b_l)
\end{math}
denote the submatrix of $\M{A}$ contained in the rows indexed by $a_1$, $a_2$, \dots, $a_{k-1}$, $a_k$ and the columns indexed by $b_1$, $b_2$, \dots, $b_{l-1}$, $b_l$.  With this notation, the condition that all the $2 \times 2$ minors of $\M{J}$ should vanish can be stated as
\begin{equation}
  \label{eq:20}
  \det \M{J}(a,b; c,d) = 0
  \quad
  (a,b \in \{1,2,3\}, a <
    b; 
    c, d \in \{1, \dots, 3I - 3\}, c < d).
\end{equation}

It is directly verified that if $\lambda_1, \dots \lambda_I$ are non-zero scalars, then \[\M{J}_i(\lambda_1 \M{H}_1, \lambda_i \M{H}_i) = \lambda_i \M{J}_i(\M{H}_1, \M{H}_i)\] for each $i = 2, \dots, I$.  This implies that
\begin{equation}
  \label{eq:21}
  [\det \M{J}(a,b; c,d)]( \lambda_1   \M{H}_1, \dots, \lambda_i   \M{H}_I)  
  =
  \lambda_{i_c} \lambda_{i_d}
  [\det \M{J}(a,b; c,d) ](\M{H}_1,  \dots,  \M{H}_I) ,
\end{equation}
where the indices $i_c$ and $i_d$ are such that the $c$-th column of $\M{J}$ belongs to $\M{J}_{i_c}$ and the $d$-th column of $\M{J}$ belongs to $\M{J}_{i_d}$ ($i_c, i_d \in \{2, \dots, I \}$), respectively.  The above identity reveals that the vanishing of
 \begin{math}
   [\det \M{J}(a,b; c,d)]( \lambda_1   \M{H}_1, \dots, \lambda_i   \M{H}_I)  
 \end{math}
 is equivalent to the vanishing of
\begin{math}
  [\det \M{J}(a,b; c,d) ](\M{H}_1, \dots, \M{H}_I).
\end{math}
Thus equations \eqref{eq:20} are genuine constraints on the homographies represented by the matrices $\M{H}_i$.

Another consequence of \eqref{eq:21} is that, being scale dependent, the functions
\begin{math}
  (\M{H}_1,   \dots,  \M{H}_I)  \mapsto [\det \M{J}(a,b; c,d) ](\M{H}_1,   \dots,  \M{H}_I)  
\end{math}
cannot be directly used as building blocks for a measure qualifying the extent to which members of a given set of $I$ homographies are mutually incompatible.  Instead, these functions have to be replaced by their scale-invariant counterparts given by
\begin{equation}
  \label{eq:22}
  \phi_{abcd}(\M{H}_1,   \dots,  \M{H}_I)
  =
 \frob{\M{H}_{i_c}}^{-1} \frob{\M{H}_{i_d}}^{-1}
  [\det \M{J}(a,b; c,d) ](\M{H}_1,   \dots,  \M{H}_I).
\end{equation}
Here, for a given matrix $\M{A}$, $\frob{\M{A}}$ denotes the Frobenius norm of $\M{A}$.  Strictly speaking, the functions $\phi_{abcd}$ are positive scale independent---their sign may still change with a change of the scales of the homography matrices.  However, the squares of these functions are genuinely scale invariant.  With this in mind, a natural measure for assessing the amount of incompatibility amongst a set of $I$ homographies can be defined by the expression
\begin{equation}
  \label{eq:23}
  \psi
  \colonequals
  \sum_{%
    \substack{%
      a,b \in \{1,2,3\}, a <
      b; \\
      c, d \in \{1, \dots, 3I - 3\}, c < d
    }}%
  \phi_{abcd}^2.
\end{equation}
It is obvious that the constraints given in \eqref{eq:20}, or, equivalently, all the constraints of the form $\phi_{abcd} = 0$, are satisfied if and only if
\begin{equation}
  \label{eq:24}
  \psi = 0.
\end{equation}

\section{Maximum Likelihood Estimation}
\label{sec:maximum-likelihood-estimation}

Let $\{\{\vt{m}_{ij}, \vt{m}'_{ij} \}_{j=1}^{J_i}\}_{i=1}^I $ be a collection of $I$ sets of pairs of corresponding inhomogeneous points in two images, arising from $I$ planar surfaces in the 3D scene.  Suppose that homography estimates $\widehat{\M{H}}_1, \dots, \widehat{\M{H}}_I$ are to be evolved based on $\{\{\vt{m}_{ij}, \vt{m}'_{ij} \}_{j=1}^{J_i}\}_{i=1}^I$ in such a way that the aggregate $\widehat{\M{H}}_1, \dots, \widehat{\M{H}}_I$ satisfies constraints \eqref{eq:20}.  One statistically meaningful approach to this estimation problem involves the \emph{maximum likelihood} cost function, also called the \emph{reprojection error} cost function,
\begin{multline}
  \label{eq:25}
  \costML(\{\M{H}_i\}_{i=1}^I,
  \{ \{ \veot{m}_{ij}  \}_{j=1}^{J_i} \}_{i=1}^I)
  \\
  =
  \sum_{i = 1}^{I} \sum_{j=1}^{J_i}
  \left( \norm{\vt{m}_{ij}  -  \veot{m}_{ij}}^2
    +
    \norm{\vt{m}'_{ij}  - \h2c{\left( \M{H}_i
          \ctwoh{(\veot{m}_{ij})}  
        \right)}}^2 \right) 
\end{multline}
that has for its collective argument the homography matrices $\M{H}_i$ and the corrections $\veot{m}_{ij}$ of the points $\vt{m}_ {ij}$ in the first image \cite[Sect.  4.5]{Hartley2004Multiple}.  Here, $\ctwoh$ and $\h2c$ denote the operators of homogenisation and dehomogenisation given by
\begin{displaymath}
  \ctwoh{(\vt{x})} = [x_1, x_2, 1]^\T
  \quad
  \text{for $\vt{x} = [x_1, x_2]^\T$}
\end{displaymath}
and
\begin{displaymath}
  \h2c{(\ve{y})}
  =
  [y_1/y_3, y_2/y_3]^\T
  \quad
  \text{for $\ve{y} = [y_1,y_2,y_3]^\T$,}
\end{displaymath}
respectively; these operators convert between the Cartesian and homogeneous coordinate representation of a given 2D point.  Minimisation of $\costML$ subject to the constraints $\phi_{abcd} = 0$ (recall the definition given in \eqref{eq:22}) yields estimates $\widehat{\M{H}}_i$ and $\widehat{\veot{m}}_{ij}$.  The $\widehat{\veot{m}}_{ij}$ may be discarded and then the remaining $\widehat{\M{H}}_i$ constitute the \emph{gold standard} maximum likelihood homography estimates.  We remark that, with a view to easing implementation, an alternative optimisation approach can be adopted---and, in fact, we use this approach in our experiments---whereby $\costML$ is minimised subject to constraints \eqref{eq:20} and the additional constraints $\frob{\M{H}_i} = 1$ ($i = 2, \dots, I$).

\section{Experiments}
\label{sec:experiments}

\begin{figure}[!htbp] 
  \subfloat[\label{fig:4}]{%
    \includegraphics[width=0.48\textwidth]{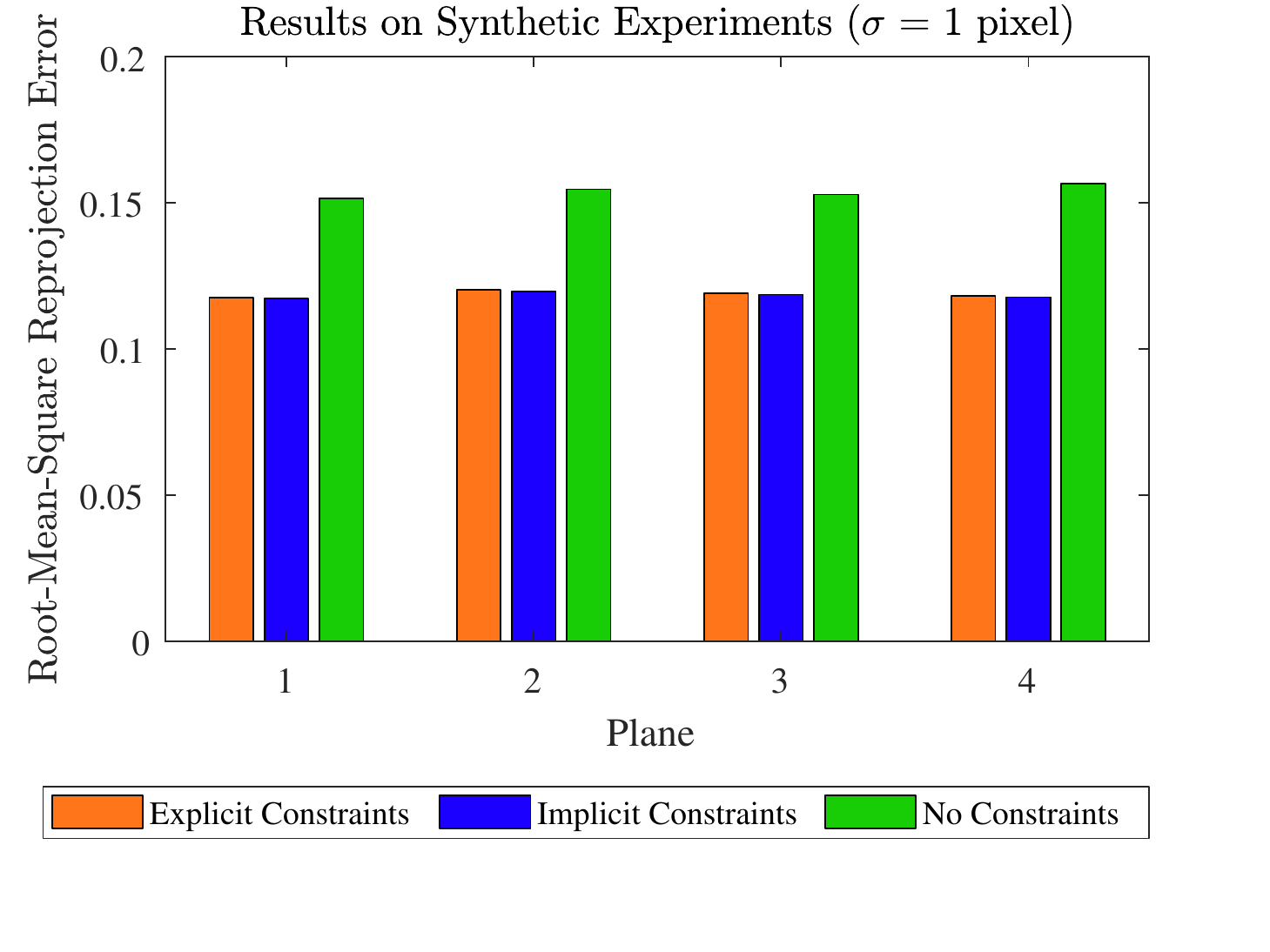} 
  } 
  \hfill 
  \subfloat[\label{fig:5}]{%
    \includegraphics[width=0.48\textwidth]{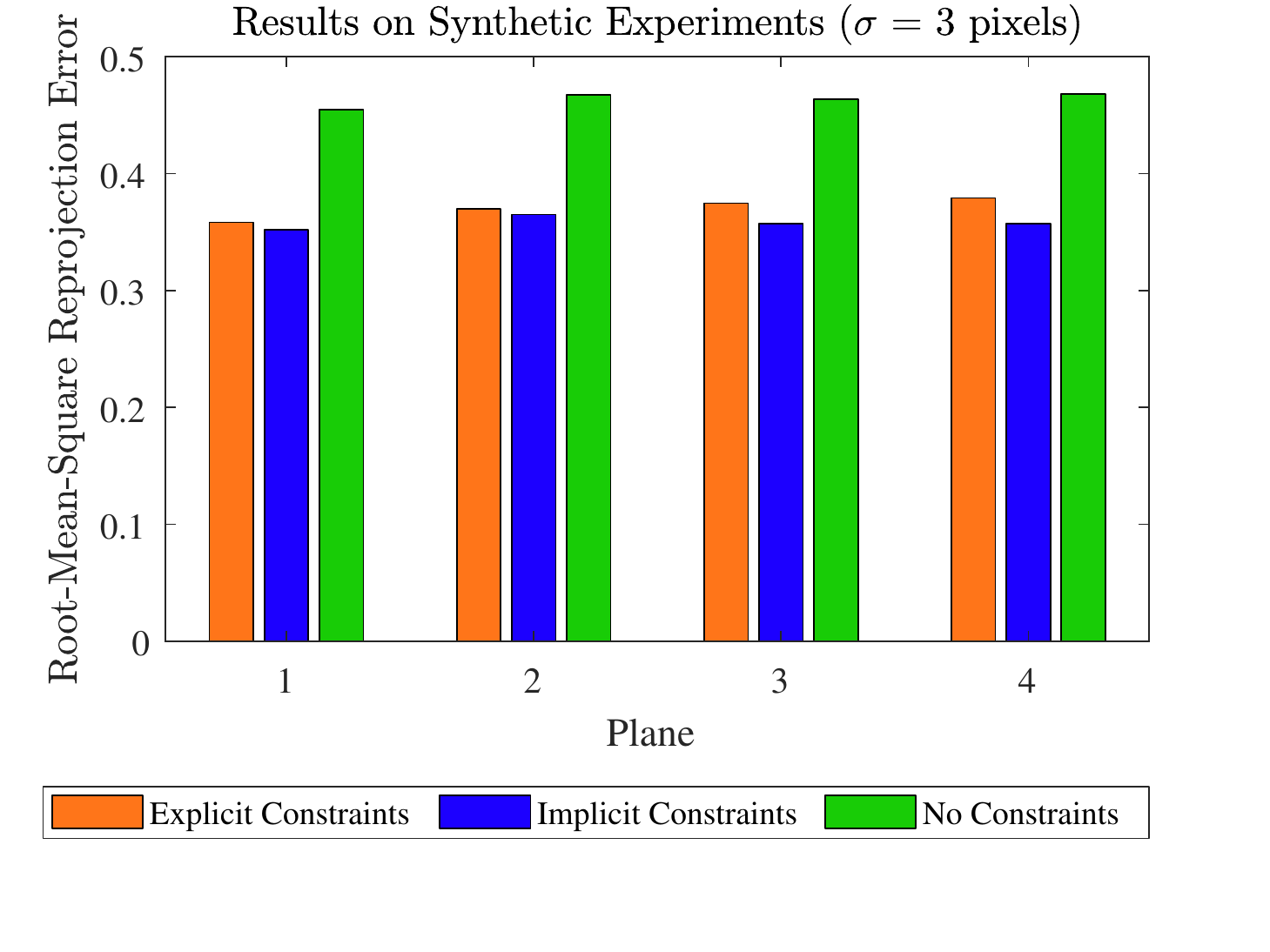} 
    \label{fig:synthetic-noise-3}   
  } 
  \caption{Experiments with simulated data evidence that enforcing full consistency constraints explicitly or implicitly improves the accuracy of the homography estimates. For each experimental trial we generated a random scene with four planar surfaces and sampled 50 corresponding points within arbitrarily sized rectangular regions. We subsequently added zero-mean isotropic Gaussian noise to the correspondences.  The accuracy of the estimates was evaluated by computing the root-mean-square reprojection error. Panels (a) and (b) show the results for Gaussian noise with a standard deviation of one and three pixels, respectively. The results are based on a thousand trials.} 
  \label{fig:synthetic}
\end{figure}

\begin{figure}[!htbp] 
\centering
  \subfloat[\label{fig:real-data-setup-experiment}]{%
    \includegraphics[width=0.65\textwidth]{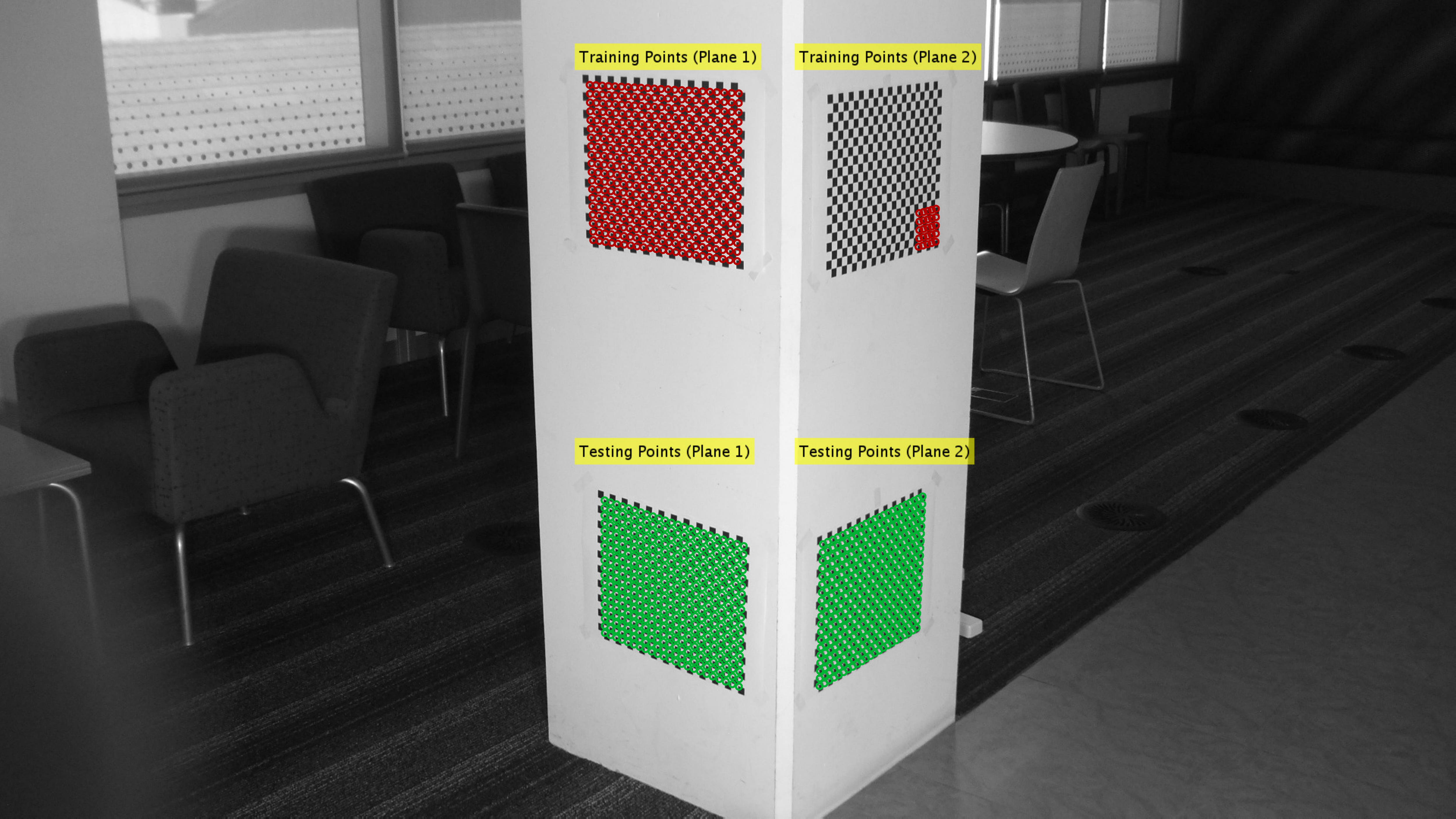} 
  }  \\
  
  \subfloat[\label{fig:real-data-plane-1}]{%
    \includegraphics[width=0.48\textwidth]{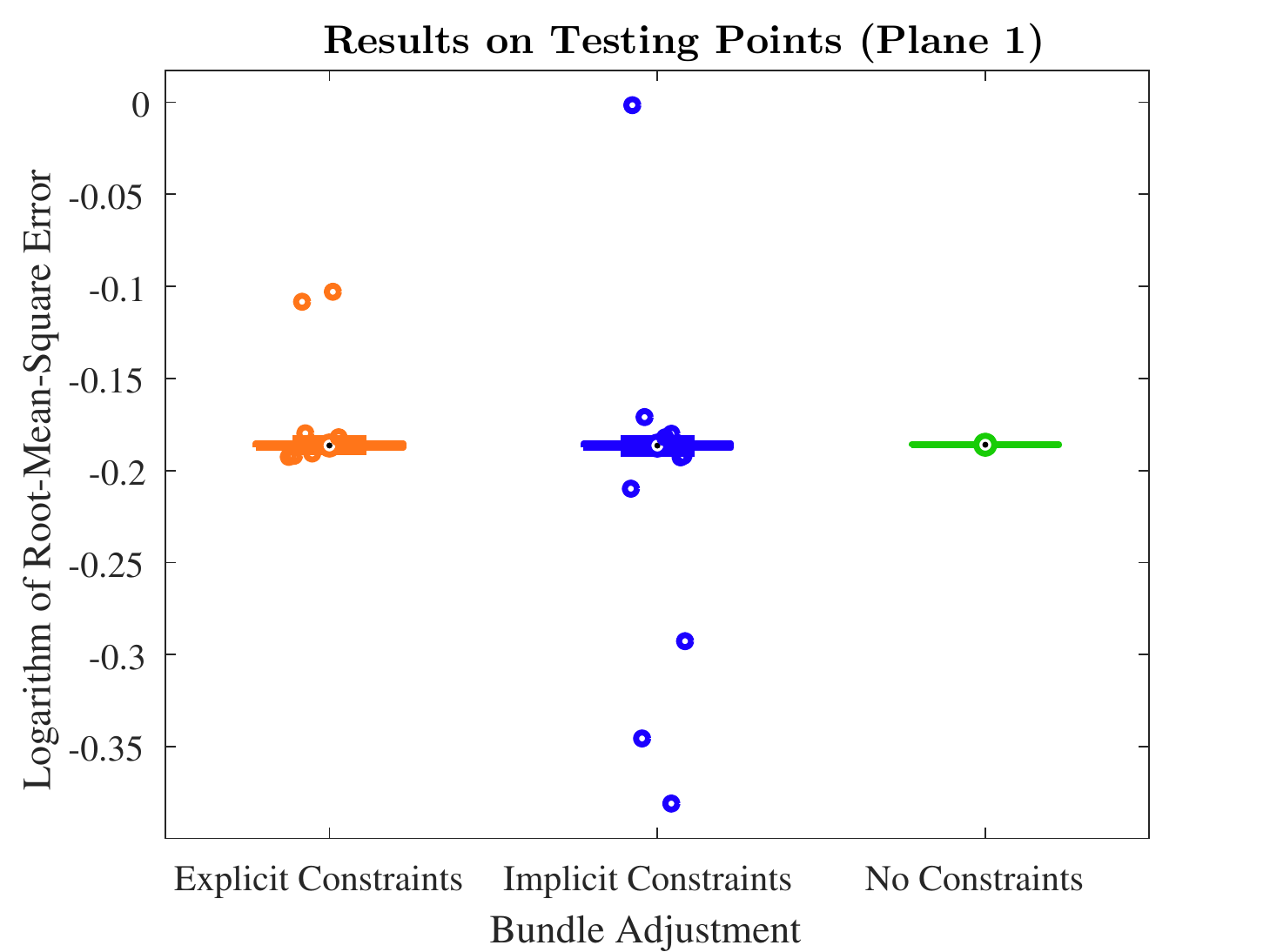} 
  } 
  \hfill 
  \subfloat[ \label{fig:real-data-plane-2}]{%
    \includegraphics[width=0.48\textwidth]{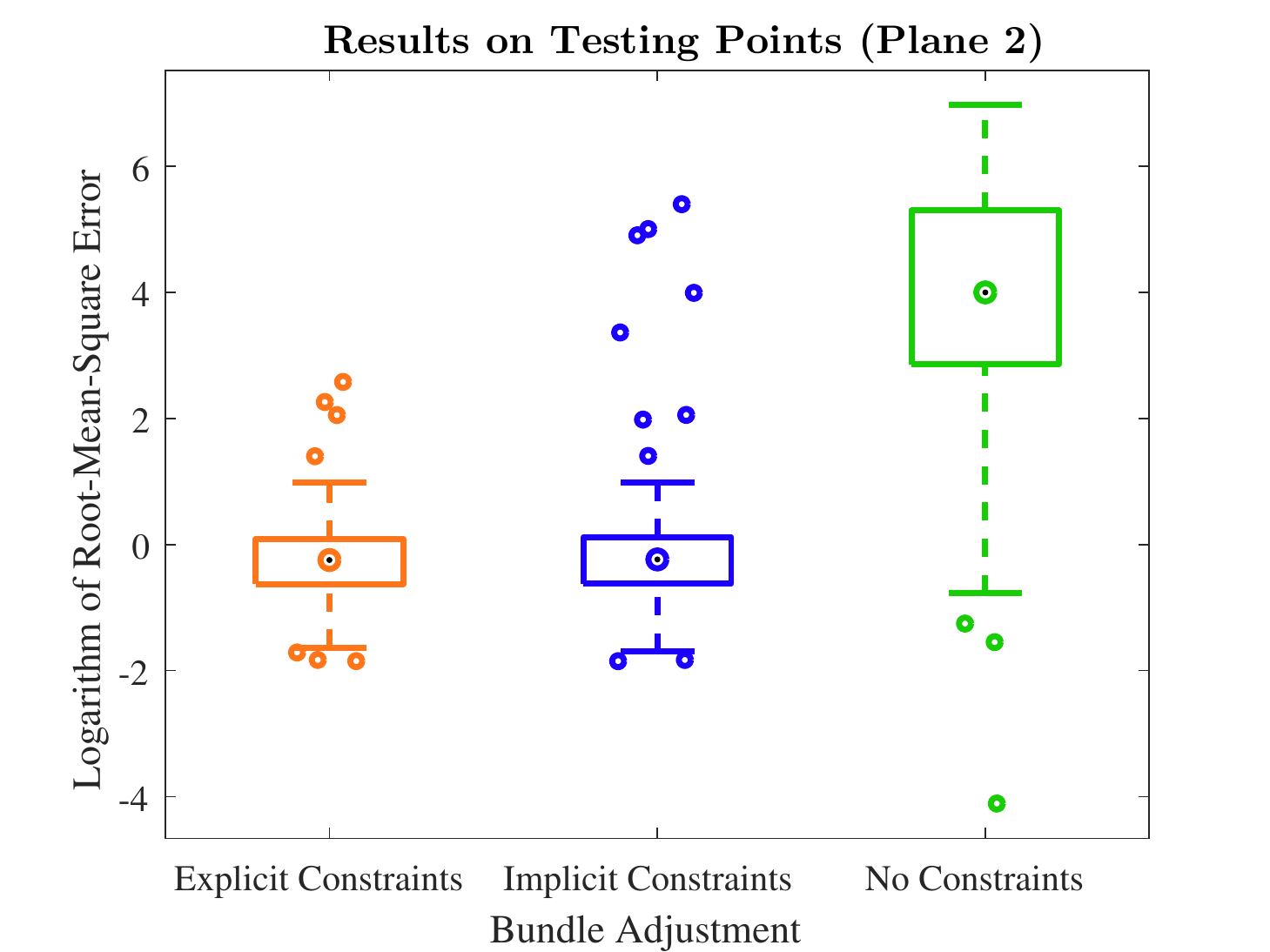} 
  } 
  \caption{Repeated experiments with realistic images demonstrate that enforcing full consistency constraints explicitly  improves the accuracy of the homography estimates. Panel (a) illustrates red feature points on two checkerboards corresponding to two different planes in three-dimensional space  (only the points in the first view are shown).  The red points indicate \emph{training data} in the sense that the points were used to estimate a pair of homography matrices (one for each plane). The green feature points on the  two other checkerboards lie on the same  planar surfaces as the training data  and served as \emph{testing data}. Note that the training data associated with the first plane spanned the entire checkerboard, whereas the training data corresponding to the second plane occupied only a small $4 \times 4$ square on the checkerboard. 
    The quality of the estimated homographies was evaluated by analysing  how accurately they mapped the testing data (green points)  into the second view. We conducted numerous experiments in which we fixed the training data for the first plane, and exhaustively varied the small $4 \times 4$ square from which the training data on the second plane were sampled.   Panels (b) and (c) show the logarithm of the root-mean-square reprojection error for the testing data in the first and second plane respectively.
    Because of the abundance of training data the estimators produced almost identical results for the first plane. However, when operating with fewer training data points, bundle adjustment without constraints produced results for the second plane which are orders of magnitude worse than variants of bundle adjustment which enforced explicit or implicit constraints. } 
  \label{fig:real-data-experiment}
\end{figure}

\begin{figure}[!htbp] 
\centering
  \subfloat[\label{fig:nese1}]{%
    \includegraphics[width=0.48\textwidth]{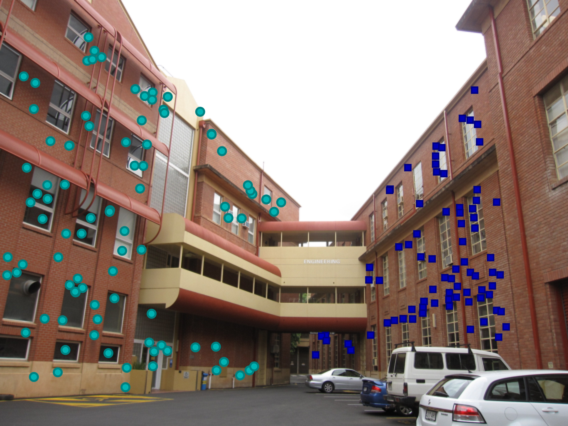} 
  } 
    \hfill 
  \subfloat[ \label{fig:nese2}]{%
    \includegraphics[width=0.48\textwidth]{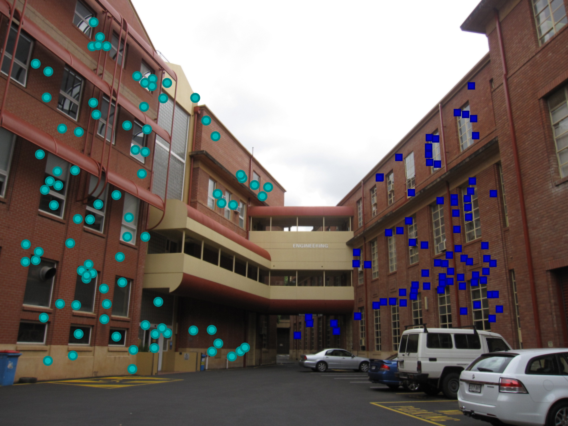} 
  } 
  \\
    \subfloat[\label{fig:library1}]{%
    \includegraphics[width=0.48\textwidth]{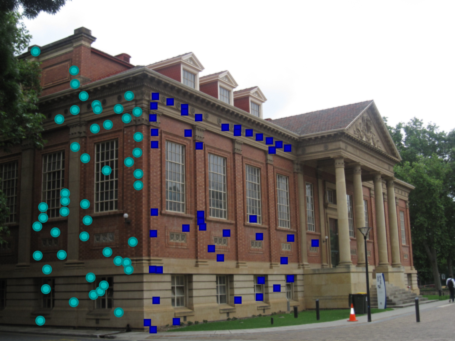} 
  } 
    \hfill 
  \subfloat[ \label{fig:library2}]{%
    \includegraphics[width=0.48\textwidth]{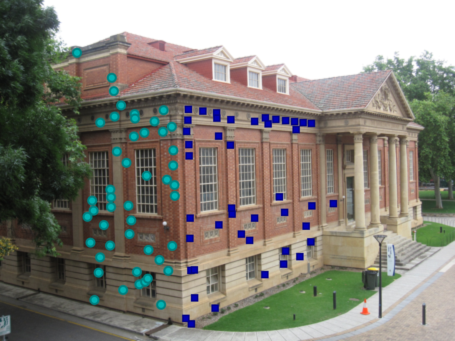} 
  } 
  \\
    \subfloat[\label{fig:library-result}]{%
    \includegraphics[width=0.48\textwidth]{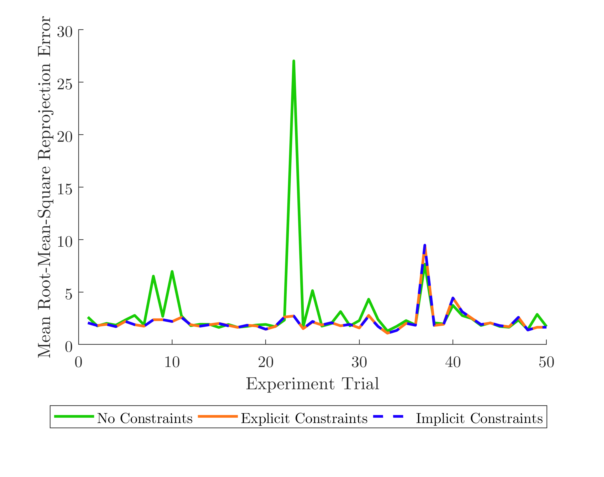} 
  } 
    \hfill 
  \subfloat[ \label{fig:nese-result}]{%
    \includegraphics[width=0.48\textwidth]{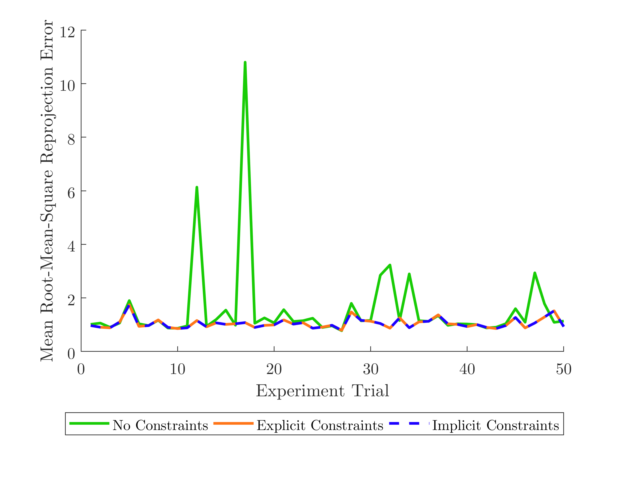} 
  } 
  \caption{Repeated experiments with images taken from the AdelaideRMF dataset
    \cite{wong:rmf}. Panels (a)--(d) illustrate manually matched SIFT feature points on two planar surfaces of the \emph{nese} and \emph{library} buildings.    For each building, we conducted 50 experiments in which we estimated a pair of homographies from ten randomly selected corresponding points. We quantified the quality of the homography estimates by evaluating the reprojection errors on the remaining set of corresponding points. Panels (e) and (f) illustrate the average root-mean-square reprojection error for the feature points in the  \emph{nese} and \emph{library} buildings, respectively.  The results of the implicit and explicit constraint enforcement algorithms are indistinguishable, and both are superior to standard bundle-adjustment. }  
  \label{fig:real-data-experiment-adelaide}
\end{figure}

We investigated the stability and accuracy of our method by conducting experiments on both synthetic and real data.  We compared our results with the gold standard bundle-adjustment method which does not enforce homography constraints \cite[Sect.  4.5]{Hartley2004Multiple}, as well as bundle-adjustment which imposes all constraints implicitly using the parametrisation outlined in \cite{szpak14:_samps}.  For all of our experiments, we ensured that there were no mismatched corresponding points. Avoiding outliers allowed us to assess the contribution of the consistency constraint enforcement on the quality of the estimated homographies by using the canonical least-squares reprojection error.  In principle, our explicit consistency constraints can also be enforced in conjunction with a robust loss function such as the Huber norm which can accommodate outliers. We estimated initial homography matrices using the direct linear transform and all estimation methods operated on Hartley-normalised data points \cite{pami:nals}.

Details on the design and outcome of our experiments with simulated and authentic image data are presented in the captions of Figures~\ref{fig:synthetic},~\ref{fig:real-data-experiment}, and~\ref{fig:real-data-experiment-adelaide}, respectively.  The results demonstrate that we have formulated a new homography estimation method capable of outperforming the established gold standard bundle-adjustment method.

The conclusions on simulated data suggest that the explicit and implicit constraint enforcement algorithms produce, on average, similar results.  The small differences between the performance of explicit versus implicit constraint enforcement in Figure~\ref{fig:synthetic-noise-3} can be attributed to the peculiarities of different optimisation schemes and the non-linear nature of the objective function. The objective function with implicit constraints
\begin{displaymath}
  (\etb, \{ \{ \veot{m}_{ij}  \}_{j=1}^{J_i} \}_{i=1}^I )
  \mapsto
 \costML(\{\Pib_i(\etb)\}_{i=1}^I,
  \{ \{ \veot{m}_{ij}  \}_{j=1}^{J_i} \}_{i=1}^I),
\end{displaymath}
with $\costML$ given in \eqref{eq:25} and $\Pib_i(\etb)$ given in \eqref{eq:4}, was optimised using the \emph{Levenberg-Marquardt} algorithm. To optimise $\costML$ subject to the explicit constraint \eqref{eq:24}, we used MATLAB's \texttt{fmincon} function set to the \texttt{interior point} algorithm. The results on authentic images are in agreement with the simulated conclusions. The experiments with real data stress the utility of imposing constraints when very few feature points are observed on one of the planar surfaces.  The logarithmic scale of the $y$-axis in Figure~\ref{fig:real-data-plane-2} shows that the homography corresponding to the second planar surface was estimated with superior accuracy. The results presented in Figures~\ref{fig:library-result} and \ref{fig:nese-result} further underscore the practical utility of the proposed algorithm. 

\section{Conclusion}
\label{sec:conclusion}

Our paper addressed a long-standing question that has evaded the research community. We have identified a complete set of constraints that need to be imposed on a set of homography matrices linking images of planar surfaces between a pair of views to ensure consistency between all the matrices. Furthermore, we have demonstrated how the constraints can be incorporated into a non-linear constrained optimisation method.  Our experiments with simulated and real images illustrated the benefits of imposing constraints in practical scenarios.

\section*{Acknowledgements}
\label{sec:acknowledgements}

This research was supported by the Australian Research Council.

\FloatBarrier

\bibliographystyle{splncs}
\bibliography{fullconstraints-bib}

\end{document}